# 4X4 Census Transform

Olivier Rukundo

This paper proposes a 4X4 Census Transform (4X4CT) to encourage further research in computer vision and visual computing. Unlike the traditional 3X3 CT which uses a nine pixels kernel, the proposed 4X4CT uses a sixteen pixels kernel with four overlapped groups of 3X3 kernel size. In each overlapping group, a reference input pixel profits from its nearest eight pixels to produce an eight bits binary string convertible to a grayscale integer of the 4X4CT's output pixel. Preliminary experiments demonstrated more image textural crispness and contrast than the CT as well as alternativeness to enable meaningful solutions to be achieved.

*Introduction:* The CT was introduced to become a very important approach to the visual correspondence problem in Computer vision and beyond [1], [2]. It basically uses a 3X3 neighbourhood or kernel size for every input pixel to generate 8 bits binary string convertible to a grayscale integer of the corresponding CT's output pixel [3]. More details/discussions on CT and its new versions and/or applications are extensively available in the literature, therefore not included in this paper. However, current literature on CT demonstrated its popularity and importance in Computer vision [4],[5], but with no particular emphasis on Visual Computing problems such as single super-resolution or resolution enhancement [6],[7],[8],[9]. In this paper, a 4X4CT is introduced to push frontiers of both computer vision and visual computing research. Unlike the traditional CT (or 3X3CT) which uses a nine pixels kernel, the proposed 4X4CT uses a sixteen pixels kernel. In such a kernel, there are four 3X3 kernel size overlapped groups. In each overlapping group, a reference input pixel uses its nearest eight pixels to achieve an eight bits binary string convertible to a grayscale integer of the corresponding 4X4CT's output pixel. The next part of this paper will focus on the proposed 4X4CT. The experimental part will present images generated using the CT and 4X4CT. The conclusion is given in the last part of the paper.

*The proposed 4X4CT:* The proposed 4X4CT runs on four overlapped groups of 3X3 kernel size, each, as shown in Figure 1. In such groups (or simply entire 4X4 kernel), all pixels are alternatively used as reference input pixels.

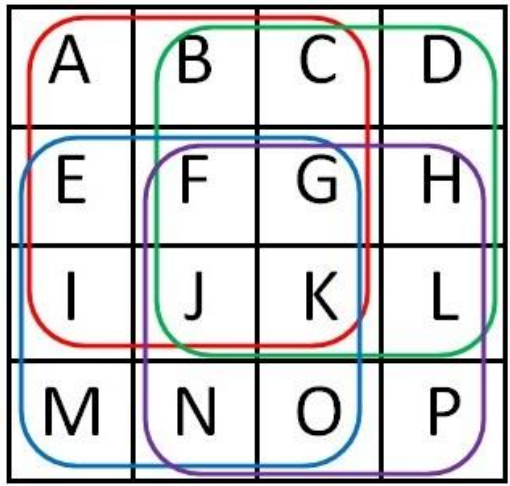

*Figure 1: The letters A to P represent reference pixels. Four overlapping pixels groups are shown in red, green, blue and purple rectangles.*

Therefore, each reference input pixel searches - in the direction indicated in Figure 2 - and uses its nearest eight pixels belong to the same overlapping group- as shown in Figure 1 and Figure 3 - to produce eight bits binary string convertible to a grayscale integer of the 4X4CT's output pixel.

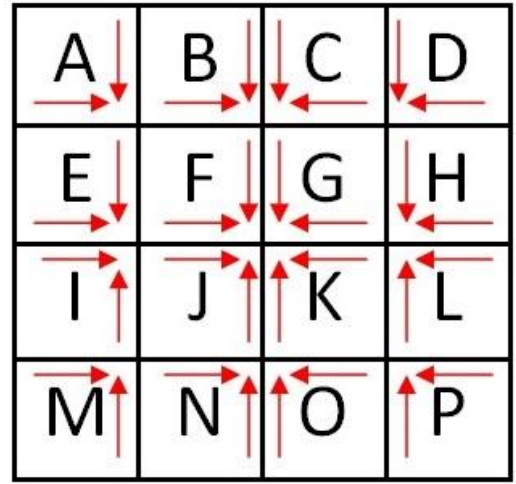

*Figure 2: Each reference pixel searches for the nearest eight pixels in the overlapping group extending over the directions indicated by the arrows in red.*

In such a binary string produced, the direction of bit position directly from the most significant bit to the least significant bit is indicated in Figure 3 by the dashed-red line.

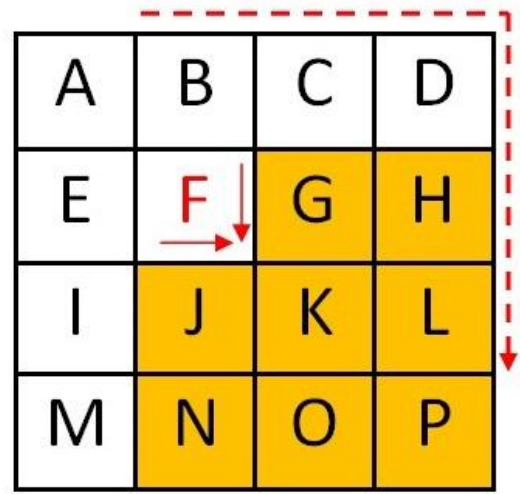

*Figure 3: The dashed red line indicates the bit position direction for all binary strings produced. The yellow area shows the overlapping group area used to produce the new integer value of F.*

Based on the Figure 3 example, if the gray level or value of the reference pixel *F* is greater than G, zero bit is recorded otherwise one bit is recorded; until an 8 bits string is formed according to the direction indicated by the dashed red line. Repeating the same operation with all other reference input pixels, composing the 4X4 kernel, leads to sixteen strings of eight bits each. Such strings are converted into grayscale integers representing the corresponding 4X4CT's output pixels. For a numerical example, suppose that the reference pixel F (in Figure 3) is equal to 4 as shown in Table 1. After the 4X4CT conversion, the new F value becomes 14. More details on how the 4X4CT worked for every reference input pixel (to achieve 4X4CT's output pixels) are provided in Table 2.

*Table 1: 4X4 input and output arrays with grayscale levels varying conventionally between 0 and 255*

| INPUT | | | | 4X4CT OUTPUT | | | |
|---|---|---|---|---|---|---|---|
| 16 | 2 | 3 | 13 | 0 | 255 | 191 | 0 |
| 5 | *4* | 10 | 8 | 251 | *14* | 67 | 206 |
| 9 | 7 | 6 | 2 | 140 | 61 | 143 | 32 |
| 4 | 14 | 15 | 1 | 255 | 2 | 0 | 255 |

*Table 2: The full process of conversion from reference input to the corresponding 4X4CT output.*

| INPUT | 4X4CT CONVERSION | | | | | | | | OUT |
|---|---|---|---|---|---|---|---|---|---|
| 0 if 16> <br> 1 if 16< | 2 | 3 | 5 | 11 | 10 | 9 | 7 | 6 | |
| **16** | **0** | **0** | **0** | **0** | **0** | **0** | **0** | **0** | **0** |
| 0 if 2> <br> 1 if 2< | 3 | 13 | 11 | 10 | 8 | 7 | 6 | 12 | |
| **2** | **1** | **1** | **1** | **1** | **1** | **1** | **1** | **1** | **255** |
| 0 if 3> <br> 1 if 3< | 16 | 2 | 5 | 4 | 10 | 9 | 7 | 6 | |
| **3** | **1** | **0** | **1** | **1** | **1** | **1** | **1** | **1** | **191** |
| 0 if 13> <br> 1 if 13< | 2 | 3 | 11 | 10 | 8 | 7 | 6 | 12 | |
| **13** | **0** | **0** | **0** | **0** | **0** | **0** | **0** | **0** | **0** |

| INPUT | 4X4CT CONVERSION | | | | | | | | OUT |
|---|---|---|---|---|---|---|---|---|---|
| 0 if 5> <br> 1 if 5< | 11 | 10 | 9 | 7 | 6 | 4 | 14 | 15 | |
| **5** | **1** | **1** | **1** | **1** | **1** | **0** | **1** | **1** | **251** |
| 0 if 11> <br> 1 if 11< | 10 | 8 | 7 | 6 | 12 | 14 | 15 | 1 | |
| **11** | **0** | **0** | **0** | **0** | **1** | **1** | **1** | **0** | **14** |
| 0 if 10> <br> 1 if 10< | 5 | 11 | 9 | 7 | 6 | 4 | 14 | 15 | |
| **10** | **0** | **1** | **0** | **0** | **0** | **0** | **1** | **1** | **67** |
| 0 if 8> <br> 1 if 8< | 11 | 10 | 7 | 6 | 12 | 14 | 15 | 1 | |
| **8** | **1** | **1** | **0** | **0** | **1** | **1** | **1** | **0** | **206** |

| INPUT | 4X4CT CONVERSION | | | | | | | | OUT |
|---|---|---|---|---|---|---|---|---|---|
| 0 if 9> <br> 1 if 9< | 16 | 2 | 3 | 5 | 11 | 10 | 7 | 6 | |
| **9** | **1** | **0** | **0** | **0** | **1** | **1** | **0** | **0** | **140** |
| 0 if 7> <br> 1 if 7< | 2 | 3 | 13 | 11 | 10 | 8 | 6 | 12 | |
| **7** | **0** | **0** | **1** | **1** | **1** | **1** | **0** | **1** | **61** |
| 0 if 6> <br> 1 if 6< | 16 | 2 | 3 | 5 | 11 | 10 | 9 | 7 | |
| **6** | **1** | **0** | **0** | **0** | **1** | **1** | **1** | **1** | **143** |
| 0 if 12> <br> 1 if 12< | 2 | 3 | 13 | 11 | 10 | 8 | 7 | 6 | |
| **12** | **0** | **0** | **1** | **0** | **0** | **0** | **0** | **0** | **32** |

| INPUT | 4X4CT CONVERSION | | | | | | | | OUT |
|---|---|---|---|---|---|---|---|---|---|
| 0 if 4> <br> 1 if 4< | 5 | 11 | 10 | 9 | 7 | 6 | 14 | 15 | |
| **4** | **1** | **1** | **1** | **1** | **1** | **1** | **1** | **1** | **255** |
| 0 if 14> <br> 1 if 14< | 11 | 10 | 8 | 7 | 6 | 12 | 15 | 1 | |
| **14** | **0** | **0** | **0** | **0** | **0** | **0** | **1** | **0** | **2** |
| 0 if 15> <br> 1 if 15< | 5 | 11 | 10 | 9 | 7 | 6 | 4 | 14 | |
| **15** | **0** | **0** | **0** | **0** | **0** | **0** | **0** | **0** | **0** |
| 0 if 1> <br> 1 if 1< | 11 | 10 | 8 | 7 | 6 | 12 | 14 | 15 | |
| **1** | **1** | **1** | **1** | **1** | **1** | **1** | **1** | **1** | **255** |

*Experimental demonstrations:* As can be seen, in Figure 5, the CT output image is not crisper than 4X4 output image. In addition, the CT output image looks much noisier and unclearer than the 4X4CT image in Figure 6. Also, the CT image features were much less similar to those of the input image shown in Figure 4.

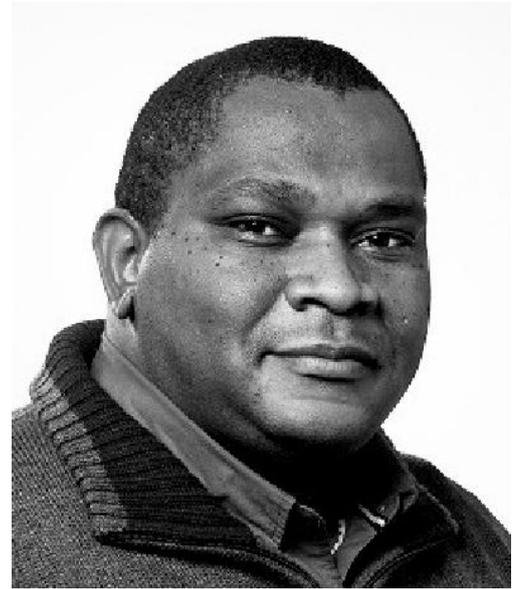

*Figure 4: Input image*

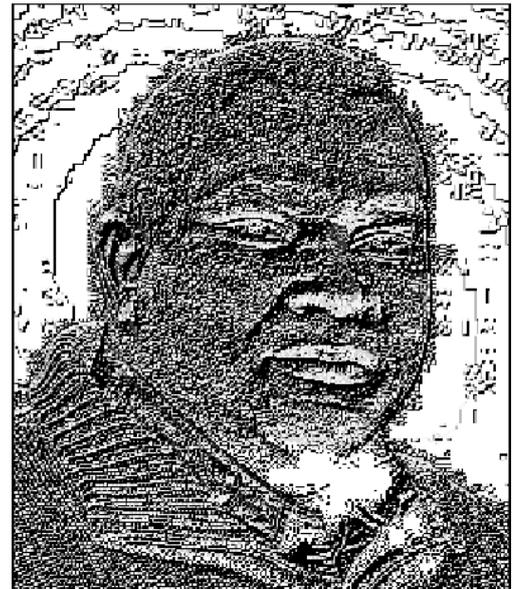

*Figure 5: CT output image*

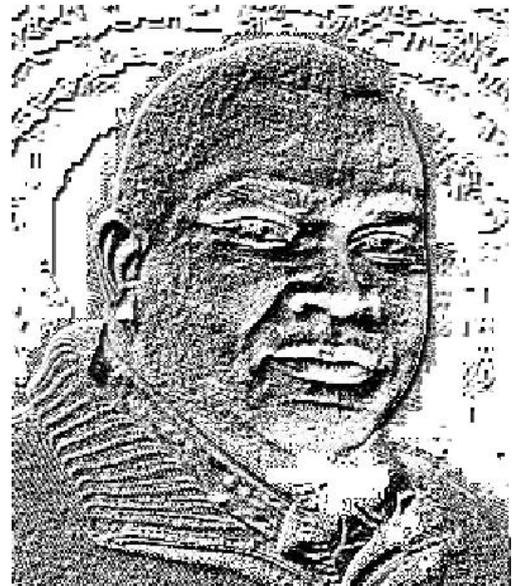

*Figure 6: 4X4CT output image*

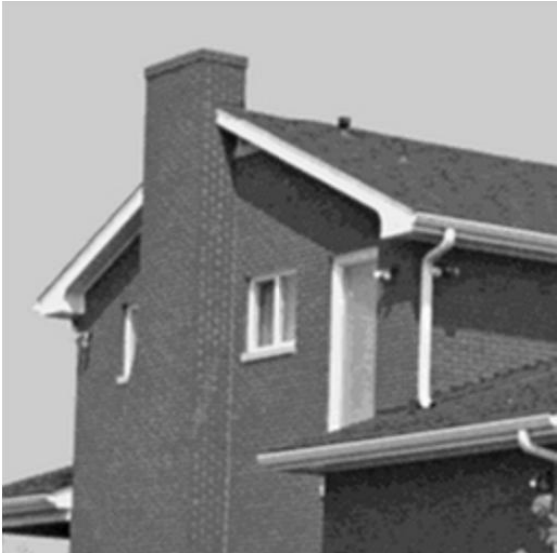
*Figure 7: Input image*

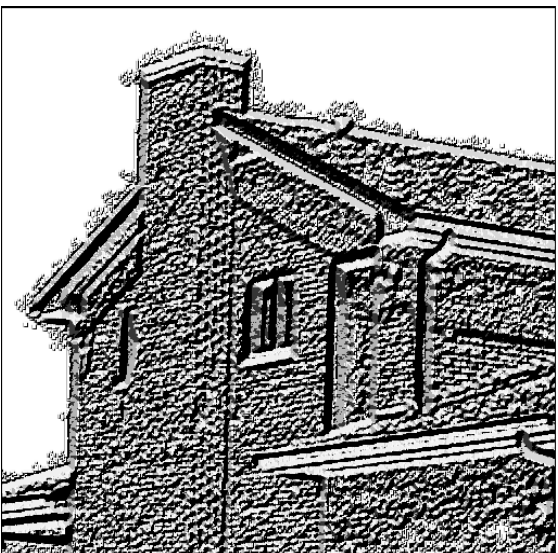
*Figure 8: CT output image*

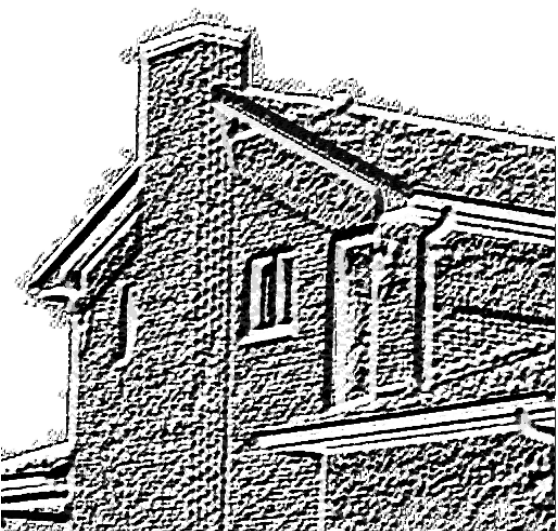
*Figure 9: 4X4CT output image*

Both images showed almost the same level of noise in the input image background. Again, in Figure 8, it is can be seen that the CT output image remained less crisp than the 4X4 output image in Figure 9.

In addition, the CT output image demonstrated lower contract in the entire output image texture than that of the 4X4CT. However, referring to the image shown in Figure 7, the 4X4CT output image features were clearer than those of the CT output image. 4X4CT demonstrated to be more image contrast and acceptable overall appearance productive. The background noise remained almost at the same level. The 4X4CT Matlab lines reading speed was slightly increased in the case involving the 4X4CT which was not an alarming case since it would be fast enough once implemented in combination with a hardware system for real-world use.

*Conclusion:* A 4X4CT was introduced and proposed in this paper, successfully. Preliminary experiments demonstrated that the 4X4CT was more promising in terms of image textural crispness and contrast than the traditional CT. They both failed equally at the background noise level, which can later be dealt with quickly by future updated or modified versions. Although current data demonstrated the alternativeness to the traditional method and encouragement to further research in computer vision and visual computing, there remains the need for further experiments for real-life solutions to be achieved.

*Acknowledgments:* This work was not supported externally.

O. Rukundo, E-mail: orukundo@gmail.com